\documentclass[authoryear]{elsarticle}
\usepackage{amsmath}
\usepackage{tikz}
\usetikzlibrary{positioning}
\begin{document}

\begin{frontmatter}
\title{Smoothing parameter estimation framework for IBM word alignment models}
\author[a]{Vuong Van Bui}
\ead{vuongbv_56@vnu.edu.vn}
\author[b]{Anh-Cuong Le \corref{cor1}}
\ead{leanhcuong@tdt.edu.vn}
\cortext[cor1]{Corresponding author}
\address[a]{VNU University of Engineering and Technology, Hanoi, Vietnam}
\address[b]{Faculty of Information Technology, Ton Duc Thang University, HoChiMinh city, Vietnam}

\begin{abstract}
  IBM models are important word alignment models in Machine Translation. 
  Based on the Maximum Likelihood Estimation principle to estimate their parameters, the models could easily overfit training data when data are sparse.
  Even though smoothing is a very popular solution in Language Model, there is still a lack of studies on smoothing for word alignment.
  In this paper, we propose a framework which generalizes the notable work \cite{moore2004improving} of applying additive smoothing to word alignment models. The framework allows developers to customize the smoothing amount for each pair of words. The added amount will be scaled appropriately by a common factor which reflects how much the framework trusts the adding strategy according to the performance on data. We also carefully examine various performance criteria and propose a smoothened version of the error count, which generally gives the best result.
  
\end{abstract}
\end{frontmatter}

\section{Introduction}
\begin{figure}
\centering
\begin{tikzpicture}[node distance=.1cm,
block/.style={
draw,
fill=white,
rectangle, 
text height={height("problems")+2pt},
font=\small}]
\node (I) [block] at (0,0) {I};
\node (do) [block, right=of I] {do};
\node (not) [block, right=of do] {not};
\node (go) [block, right=of not] {go};
\node (to) [block, right=of go] {to};
\node (the) [block, right=of to] {the};
\node (house) [block, right=of the] {house};

\node (ich) [block] at (.5,1) {ich};
\node (gehe) [block, right=of ich] {gehe};
\node (ja) [block, right=of gehe] {ja};
\node (nicht) [block, right=of ja] {nicht};
\node (zum) [block, right=of nicht] {zum};
\node (haus) [block, right=of zum] {haus};

\draw[->] (I) -- (ich);
\draw[->] (not) -- (nicht);
\draw[->] (go) -- (gehe);
\draw[->] (to) -- (zum);
\draw[->] (the) -- (zum);
\draw[->] (house) -- (haus);
\end{tikzpicture}
\caption{An example of word alignment} 
\label{fig:example}
\end{figure}
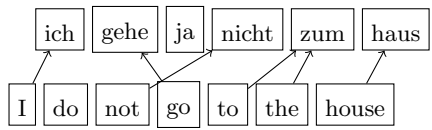

Word alignment is one of critical components in Statistical Machine Translation, which is an important problem in Natural Language Processing.
The main function of word alignment is to express the correspondence between words of a bilingual sentence pairs. In each alignment, there is a set of links whose two end-points are two words of different sides of the sentence pair. When there is a link between a pair of words, they are considered to be the translation of each other.
This kind of correspondence is usually unknown most of the time and it is usually derived from a bilingual corpus with the support of word alignment models. An example of word alignment for a sentence pair of English-German is shown in Figure~\ref{fig:example}.

IBM Models presented in \cite{brown1993mathematics} are currently the most popular word alignment models. 
Based on the Maximum Likelihood Estimation principle, the parameters of IBM Models are estimated from training data with the Expectation Maximization algorithm presented in \cite{dempster1977maximum}. This specialized algorithm is applied to determine the local maximum likelihood estimation by considering the word alignment unobserved variables of training data.

However, \emph{overfitting}, a common problem in Machine Learning, occurs in word alignment models frequently. This issue appears when the model fits the training data too well but performs poorly on the testing data. The Maximum Likelihood estimation makes the parameters ``agree'' as much as possible with the training data and nothing else. This is not always appropriate, especially when there are not sufficient training data to obtain a reliable estimation. Even thousands of sentence pairs could still contain many rare structures due to the complexity of the languages. Therefore, the right solution should not rely completely on the training data. Before observing the data, we normally do have some prior knowledge of the languages. Integrating these features usually help reducing the problems caused by sparse data.

Many methods have been developed to deal with sparse data. Bilingual dictionaries together with many other sorts of linguistic analysis such as morphological analysis, syntactic analysis have been well investigated in \cite{brown1993but}, \cite{koehn2007factored} ,\cite{sadat2006combination} and \cite{lee2004morphological}. 
Although these approaches have many good behaviors in experiments, applying one known method of a language pair for another language pair is usually difficult due to its language dependencies. 

Language Model, which is another well known problem in Machine Translation, has sparse data as the main issue to deal with. The task of Language Model is to estimate how likely a sentence is produced by speakers in which the training data is hardly able to cover all cases. A method called ``smoothing'' is a very popular solution to the issue \cite{chen1999empirical}, \cite{goodman2001bit}. The idea is that, when estimating a probability, we will give a little mass to events that do not occur in the training data.
Although the smoothened model is not the strongest one to ``agree'' with the training data, it will perform better on the testing data than the un-smoothened model.

In spite of being a very popular technique to deal with rare events, there is still lack of studies of applying smoothing to the word alignment problem. To the best of our knowledge, the only study on this subject is the work by Moore \cite{moore2004improving}. A basic additive smoothing was utilized for the parameter of word translation probability to yield a significant improvement of the model. In this paper, we will present an extended study on this technique.

Our main contribution is a new general framework that generalizes Moore's method. We first examine the additive smoothing under the view of Maximum a Posteriori estimation and figure out that it corresponds to a special case of the prior Dirichlet distribution. By utilizing the general form of the prior Dirichlet distribution, we have a general framework in which the Moore's method is just an instance of the framework. There is one difficulty that the prior distributions which are usually specified by heuristics methods may be inappropriate for many instances of training data. Therefore, we propose a method to scale the overall degree of how we trust the prior distribution. The degree of belief is learnt from the development data. Inappropriate distributions will get a low degree of belief. That means that, in the worst case, the framework still performs as well as the baseline model, while in other cases, we may achieve better results.

Besides the general framework, we carefully study parameter estimation approaches, particularly learning the degree of belief. Many objective functions are empirically evaluated to learn this parameter. They are the likelihood of sentence pairs and the likelihood of sentence pairs with their alignment and the error count. The error count appears to have the highest correlation with the evaluation metric AER (Alignment Error Rate). However, it is a discrete function of the parameter which may reduce the performance of the optimizing algorithms. Therefore, we develop a continuous approximation of the error count. As expected, this objective function gives the best overall performance.

The structure of our paper is organized as follows. After describing the related work, we present the formulations of IBM models, the method of estimating the parameters of the models together with a discussion on the problems of the estimating approach. The next section describes the Moore's basic method of additive smoothing. Then, we present our proposed framework, the approaches to estimate the parameters of the framework. The final section contains our empirical results of the methods with the discussion and conclusion.

\section{Related work}
The problem of sparsity is well known in the Machine Learning field. For word alignment, the instance of the rare word problem is studied in \cite{brown1993but} and \cite{moore2004improving}. In these papers, rare words act as ``garbage collectors'' that tend to align to too many target words.

To deal with rare word problems, many researches utilized the linguistic information. One of the earliest works is \cite{brown1993but} which used an external dictionary to improve the word alignment models. Experiments show that this method also solves the problem of rare words. Another approach utilized the information provided by morphological analysis. Some of them are \cite{koehn2007factored}, \cite{sadat2006combination}, \cite{lee2004morphological}. These works do not treat word as the smallest unit of translation. Instead, they computed statistics on morphemes, which are smaller parts of constructing words. This helps reducing sparsity by having better statistics of the rare words which are composed of popular morphemes as popular words. However, when a word is really rare, in which its morphemes are rare as well, these methods are no longer applicable. Furthermore, the dependencies on the languages also limits the scope of these methods.

The problem of rare words is also reduced with methods involving word class. IBM Model 4 (\cite{brown1993mathematics}) constrains word distortion on the class of source words. Distortion indicates how likely two words are translations of each other based on theirs positions. A better distortion estimation would result in a better alignment. Another work from \cite{toutanova2002extensions} utilized the word classes of both source words and target words. It estimated the translation probability of pairs of classes as well as the word translation probability. This class translation probability is usually more reliable than the word translation probability. Aligning will be better, especially in the case of rare words when it encourages alignments to follow the class translation probability. Word classes may be part-of-speech tags which is usually obtained by running part-of-speech tagging software (such as the Stanford tagger in \cite{toutanova2003feature}) or more usually the classes obtained by running a clustering algorithm which is language independent as in \cite{och1999efficient}.

Smoothing is a popular technique to solve the problem of sparsity. Language Model, which is another problem of Machine Translation, has a variety of smoothing methods. The classic paper \cite{chen1999empirical} gives a very good study of this solution for the Language Model problem. A large number of smoothing methods with extensive comparisons amongst them will be analyzed carefully in the paper.

However, there is still lack of studies for applying smoothing techniques to the word alignment problem. The work of this paper is mostly an extended work of \cite{moore2004improving} which is the earliest study for this matter. In that work, Moore has an intensive study of many improvements to IBM Model 1. Additive smoothing is the first of the three improvements, which directly attempts to solve the sparsity problem. In spite of the simplicity, the method has good performance. 
The smoothing technique Moore applied is additive smoothing. For every pair of word $(e, f)$, it is assumed that $e$ and $f$ were paired in a constant number of times before the training data is observed. Therefore, in each estimation, this assumption will be taken into account and the prior number of times, which they are paired, will be added appropriately. The constant prior count is tuned manually or learnt from some additional annotated data.

In this paper, we propose a general smoothing framework. This could be applied to every pair of languages because it is language independent. It also has advantages over the Moore's method when it does not force identical additive amount for each word pair. Instead, it allows developers to customize the amount based on their own strategies. These specified amounts is scaled according to the appropriateness of the strategy before being added to the counts. The scaling degree is very close to 0 when the strategy is inappropriate.  When the strategy is adding a constant amount, this instance of framework is identical to the Moore's method. Therefore, we not only have a more general framework but also have a framework with the certain that it would hardly decrease the overall results due to the scaling factor.



\section{Formal Description of IBM models}
\subsection{Introduction of IBM Models}
IBM models are very popular among word alignment models. In these models, each word in the target sentence is assumed to be the translation of a word in the source sentence. In case, the target word is not the translation of any word in the source sentence, it is assumed to be the translation of a hidden word ``NULL'' appearing in every source sentence at position 0 as convention.

There are 5 versions of IBM Models from IBM Model 1 to IBM Model 5. Each model is an extension from the previous model with an introduction of new parameters. IBM Model 1 is the first and also the simplest model with only one parameter. However, this parameter, the word translation probability which is the most important parameter is also used in all later models. A better estimation for this parameter will results in better later models. Because this paper only employs this parameter, only related specifications of IBM Model 1 is briefly discussed. More details of explanations, proofs and the later models can be found in many other sources such as the classic paper \cite{brown1993mathematics}

\subsection{IBM Model 1 formulation}
This section briefly presents the formal description of IBM model 1. With a source sentence $\textbf{e}$ of length $l$ which contains words $e_1, e_2, \ldots, e_l$ and the target sentence $\textbf{f}$ of length $m$ which  contains words $f_1, f_2, \ldots f_m$, the word alignment $\textbf{a}$ is represented as a vector of length $m$: $<a_1, a_2, \ldots, a_m>$, in which each $a_j$ means that there is an alignment link from $e_{a_j}$ to $f_j$. The model with the model parameter $\textbf{t}$ states that with the given source sentence $\textbf{e}$, the probability that the translation in the target language is the sentence $\textbf{f}$ with the word alignment $\textbf{a}$ is:

\begin{equation} \label{eq:model1gen}
  p(\textbf{f}, \textbf{a} \mid \textbf{e}; \textbf{t}) = \frac{\epsilon}{(l+1)^m} \prod_{j=1}^{m} t(f_j \mid e_{a_j})
\end{equation}

The constant $\epsilon$ is used to normalize the proper distribution. It guarantees that the probabilities of all events sum up to $1$. The word translation probability $t(f \mid e)$ reflects how likely the word $f$ is the translation of the word $e$. With the above equation \ref{eq:model1gen}, the model states that with the word translation probability, the probability of the sentence $\textbf{f}$ is the translation of $\textbf{e}$ with the word alignment $\textbf{a}$ is proportional to the product of translation probabilities of all alignment links.

With the model parameter $\textbf{t}$, we are also able to deduce the probability of an alignment for a known sentence pair.
\begin{equation}
p(\textbf{a} \mid \textbf{e}, \textbf{f}; \textbf{t}) = \prod_{j=1}^{m} \frac{t(f_j \mid e_{a_j})}{\sum_{i=0}^{l} t(f_j \mid e_i)}
\end{equation}

The distribution of which word in the source sentence is aligned to the word at the position $j$ in the target sentence is also deduced by following equation:

\begin{equation} \label{eq:p-link}
  p(a_j \mid \textbf{e}, \textbf{f}; \textbf{t}) = \frac{t(f_j \mid e_{a_j})}{\sum_{i=0}^{l} t(f_j \mid e_i)}
\end{equation}

By having the above distribution of each word alignment link, the most likely word in the source sentence to be aligned to the target word $f_j$ is:

\begin{align}
  {\hat{a}}_j &= \max\arg_i p(a_j = i \mid \textbf{e}, \textbf{f}) \\
  &= \max\arg_i \frac{t(f_j \mid e_i)}{\sum_{i=0}^{n} t(f_j \mid e_i)} \\
  &= \max\arg_i t(f_j \mid e_i)
\end{align}

For each target word, the most likely correspondent word in the source sentence is the word giving the highest word translation probability for the target word among all words in the source sentence. It means that with the model, we can easily determine the most likely alignment $\hat{\textbf{a}}$ of a sentence pair, which is often known as the Viterbi alignment.

\section{Parameter Estimation, The Problems and The Current Solution}
\subsection{Estimating the parameters with Expectation Maximization algorithm}
The parameter of the word alignment model is not usually given in advance. Instead, all we have in the training data is often just a corpus of bilingual sentence pairs. Therefore, we have to determine the parameter which maximizes the likelihood of these sentence pairs. The first step is to formulate this likelihood. The likelihood of one sentence pair $(\textbf{e}, \textbf{f})$ is:
\begin{align}
  p(\textbf{f} \mid \textbf{e}; \textbf{t}) &= \sum_{\textbf{a}} p(\textbf{f}, \textbf{a} \mid \textbf{e}; \textbf{t}) \\
  &= \sum_{\textbf{a}} \frac{\epsilon}{(l+1)^m} \prod_{j=1}^{m} t(f_j \mid e_{a_j}) \\
  &= \frac{\epsilon}{(l+1)^m} \prod_{j=1}^{m} \sum_{i=0}^{l} t(f_j \mid e_i)
\end{align}

The likelihood of all pairs in the training set is the product of the likelihoods of the individual pairs with an assumption of conditional independence between pairs with the given parameter.
\begin{equation}
  p(\textbf{f}^{(1)}, \textbf{f}^{(2)}, \ldots, \textbf{f}^{(n)} \mid \textbf{e}^{(1)}, \textbf{e}^{(2)}, \ldots, \textbf{e}^{(n)}; \textbf{t}) = \prod_{k=1}^{n} p(\textbf{f}^{(k)} \mid \textbf{e}^{(k)}; \textbf{t})
\end{equation}
There is no closed-form solution for the parameter which maximizes the likelihood of all pairs. There is, instead, an iterative algorithm, Expectation Maximization algorithm (\cite{dempster1977maximum}), which is suitable for this particular kind of problem. At first, the algorithm initiates an appropriate value for the parameter. Then, the algorithm iterates to fit the parameter to the training data in term of likelihood. The algorithm stops when either the limit number of iterations reached or likelihood convergence spotted.

Each iteration consists of two phases: the Expectation phase and the Maximization phase. In the Expectation phase, it estimates the probability of all alignments using the current value of the parameter as in Equation~\ref{eq:p-link}. Later, in the Maximization phase, the probabilities of all possible alignments estimated from the current value is used to determine a better value for the parameter in the next iteration as following.


Denote $count(e, f)$ as the expected number of times the word $e$ is aligned to the word $f$, which is calculated as:
\begin{equation}
  count(e, f \mid \textbf{t}) = \sum_k \sum_j \sum_i [f = f^{(k)}_j] [e = e^{(k)}_i] p(a_j = i \mid \textbf{e}^{(k)}, \textbf{f}^{(k)}; \textbf{t})
\end{equation}

We also denote $count(e)$ as the expected number of times the word $e$ is aligned to some word in the target sentences, which is calculated in term of $count(e, f)$ as:
\begin{equation}
  count(e \mid \textbf{t}) = \sum_f count(e, f \mid \textbf{t})
\end{equation}

These counts are calculated in term of $p(a_j = i)$, which is actually calculated in term of the current parameter of the model $t(f \mid e)$ as in Equation~\ref{eq:p-link}. The model parameter for the next iteration, which is better in likelihood, is calculated in term of the current counts as:
\begin{equation}
  t_{i+1}(f \mid e) = \frac{count(e, f \mid \textbf{t}_{i})}{count(e \mid \textbf{t}_{i})}
\end{equation}

The parameter, which is estimated after each iteration $i+1$, will improve the likelihood of the training data over that of the previous iteration $t$. After getting a desired convergence, the algorithm stop and return the estimated parameter. With that parameter, it is easy to deduce the most likely alignment on new pairs of sentences.

\subsection{Problems with Maximum Likelihood Estimation}
Sparseness is a common problem in Machine Learning problems. For the word alignment problem, there is hardly an ideal corpus with a variety of words, phrases and structures which appear at high frequencies and are able to cover almost all cases of the languages. In real datasets, there would be a lot of rare events and missing cases. The popular Maximum Likelihood estimation relies completely on the training data and its estimation is usually not very reliable due to the spareness of data. 

Due to the complicated nature of languages, spareness often appears at many levels of structures such as words, phrases, etc. Each level has its own complexity and effect to the overall spareness. In this paper, we only investigate the spareness of words. 
We believe that this study could motivate further studies on the spareness of more complex structures.

In this section, the behavior of rare words will be studied. ``Rare'' words are words that occur very few times in the corpus. No matter how large the corpus is, there are usually many rare words. Some of them appear only once or twice. For purpose of explanation, we assume that a source word $e^*$ appears only once in the corpus. We denote the sentence containing $e^*$ to be $e_1, e_2, \ldots, e^*, \ldots, e_l$ and the corresponding target sentence to be $f_1, f_2, \ldots, f_m$. Due to the only occurrence, in the word translation probability of $e^*$, of all words in the target vocabulary, only $f_1, f_2, \ldots, f_m$ could have positive probabilities. All probabilities of these words sum up properly to 1 while the probability of every other word in the vocabulary is 0. We will show why these positive quantities are usually not much different.

The likelihood of the pair of the sentences in which $e^*$ occurs is:
\begin{equation}
  \prod_j \sum_i t(f_j \mid e_i) = \prod_j (t(f_j \mid e^*) + \sum_{i: e_i \neq e^*} t(f_j \mid e_i))
\end{equation}

The translation probabilities of $e^*$ can hardly significantly affect the likelihood of all sentence pairs. Therefore, we assume that the translation probabilities of all $e_i$ other than $e^*$ are known which maximizes the likelihood of all other pairs, we now need to determine the word translation probability of $e^*$ which maximizes only the likelihood of the pair containing the rare word. 
\begin{equation}
  \prod_j (t_j + c_j)
\end{equation}
for $t_j = t(f_j \mid e^*)$ as variables and $c_j$ as constants having values of $\sum_{i: e_i \neq e^*} t(f_j \mid e_i))$

There is not a closed-form solution for the $t_j$. However, the solution must make values $t_j + c_j$ as closed to each other as possible. In many cases, $c_j$ are not too far from each other. The value of $c_j$ for which $f_j$ is real translation of $e^*$ can be lower than other, but usually not too far. Therefore, $t_j$ are not too much different from each other. When the target sentence is not too long, each target word in the target sentence will get a significant value in the translation probability of $e^*$.

This leads to the situation in which even $e^*$ and $f_j$ do not have any relation, the word translation probability $t(f_j \mid e^*)$ still has a significant value. Considering the case that the source sentence contains a popular word $e_i$ which has an infrequent translation in the target sentence $f_j$, for example, $f_j$ is the translation of $e_i$ about 10\% of the times in the corpus. The estimated $t(f_j \mid e_i)$ should be around $0.1$. If  $t(f_j \mid e^*) > 0.1$, $e_i$ will no longer be aligned to $f_j$, and $e^*$ will be aligned instead if no other $e_{i'}$ has the greater translation probability than $t(f_j \mid e^*)$. This means that a wrong alignment occurs!

This situation lead to the issue that the rare source word is aligned to many words in the target sentence. This is also explained in \cite{moore2004improving} and particular examples of this behavior can be found in \cite{brown1993but}. The overfitting is reflected in the action of estimating the new parameter merely from the expected count of links between words. For the case of rare words, these counts are small and unreliable to compute statistics. 

This is the main motivation for the smoothing technique. By various techniques to give more weight in the distribution to other words that do not co-occur with the source word or by adjusting the amount of weight derived from Maximum likelihood estimation, we could get a more reasonable word translation table.

\subsection{Moore's additive smoothing solution}
Additive smoothing, which is often known as the Laplace smoothing, is a basic and fundamental technique in smoothing. Although it is considered as a poor technique in some applications like Language Model, reasonably good results for word alignment are reported in \cite{moore2004improving}.

As in the maximization step of Expectation Maximization algorithm, the maximum likelihood estimation of the word translation probability is:
\begin{equation}
  t(f \mid e) = \frac{count(f, e)}{count(e)}
\end{equation}

Employing ideas from Laplace smoothing for Language Model, \cite{moore2004improving} applied the same method by adding a constant factor $n$ to every count of a word pair $(e, f)$. By this way, the total count of $e$ increases by an amount of $|F|$ times the added factor, where $|F|$ is the size of the vocabulary of the target language. We have the parameter explicitly estimated as
\begin{equation}
  t(f \mid e) = \frac{count(f, e) + n}{count(e) + n|F|}
\end{equation}

The quantity to add $n$ is for adjusting the counts derived from the training data. After iterating through the traditional IBM Models, we have the estimation of the expected number of times a source word is aligned to a target word. There are many target words which do not have any chance of being aligned to the source word following the estimation from the training corpus. This may be inappropriate because the source word is in fact still aligned to these target words at lower rates that makes these relations not appear in the corpus. Concluding that they do not have any relation at all by 0 probabilities due to Maximum Likelihood estimation is not reasonable. Therefore, we have an assumption for every pair of words $(e, f)$ that before observing the data, we have seen $n$ times the source word $e$ and the target word $f$ are linked. The quantity $n$ is uniquely applied to every pair of words. This technique makes the distributions generally smoother, without 0 probabilities and more reasonable due to an appropriate assumption.
 

\section{Our proposed general framework}
\subsection{Bayesian view of Moore's additive smoothing}
The additive smoothing can be explained in term of Maximum a Posteriori principle. In this section, we briefly present such an explanation. Details of Maximum a Posteriori principle can be further found in chapter 5 of \cite{murphy2012machine}.

The parameter of the model has one distribution $t_e(f) = t(f \mid e)$ for each source word $e$. As Maximum a Posteriori principle, the parameter value will be chosen to maximize the posteriori $p(\textbf{t} \mid \textbf{data})$ instead of $p(\textbf{data} \mid \textbf{t})$. However, the posteriori can be actually expressed in term of the likelihood.

\begin{equation}
  p(\textbf{t} \mid \textbf{data}) \sim p(\textbf{data} \mid \textbf{t}) \; p(\textbf{t})
\end{equation}

If the prior distribution of the parameter $p(\textbf{t})$ is uniform, maximizing the likelihood will be identical to maximizing the posteriori. However, if the prior distribution is different from the uniform, we are going to prefer some parameters than the others. In such a case, the two principles: Maximum Likelihood and Maximum a Posteriori will return different parameters.

One of the most popular choice of a prior distribution is Dirichlet distribution which has the density function.
\begin{equation}
  f\left(p_{1},\ldots ,p_{K};\alpha _{1},\ldots ,\alpha _{K}\right)={\frac {1}{\mathrm {B} (\alpha )}}\prod _{i=1}^{K}p_{i}^{\alpha _{i}-1}
\end{equation}

in which, for the word alignment problem $p_1, p_2, \ldots p_K$ is the probability of all $K$ source words in the word translation probability of a source word $e$.

The Dirichlet distribution is analogous to the assumption that we have seen $\alpha_i - 1$ occurrences of the event number $i$. The $p$ that maximizes the $f(p, \alpha)$ will be $p_i = \frac{\alpha_i - 1}{\sum_i \alpha_i - 1}$. If we see additionally $c_i$ occurrences of each event number $i$ in the training data, the $p$ that maximizes the posteriori will be $p_i = \frac{\alpha_i - 1 + c_i}{\sum_i \alpha_i - 1 + c_i}$. 

We now present an interpretation of additive smoothing in term of Maximum a Posteriori estimation with the Dirichlet distribution as the prior distribution. The parameter of the model is a set of distributions, in which each distribution corresponds to a translation probability of a source word. All these prior distributions share a same parameter $\alpha$ of Dirichlet distributions. In the maximization step of the EM algorithm, an amount of $\alpha_i - 1$ will be taken into account as the number of times the event number $i$ is observed before the times we observe the event number $i$ in the training data. Therefore, a right interpretation assigns all $\alpha_i$ in additive smoothing the same value. With that configuration of parameters, we are able to achieve an equivalent model to the additive smoothing.

\subsection{Formal Description of the framework}
Adding a constant amount to the count $(e,f)$ for every $f$ does not seem very reasonable. In many situations, some count should get more than others. Therefore, the amount to add for each count should be different if necessary. This can be stated in term of the above Bayesian view that it is not necessary to set all $\alpha_i$ the same value for the Dirichlet prior distribution. In this section, we propose a general framework which allows freedom to customize the amount to add with the certain that this hardly decrease the quality of the model.

We denote $G(e, f)$ as the amount to add for each pair of word $(e, f)$. This function is usually specified manually by users. With the general function indicating the amount to add, we have a more general model, in which the parameter is estimated as:

\begin{equation}
  t(f \mid e) = \frac{count(f, e) + G(e, f)}{count(e) + \sum_f G(e, f)}
\end{equation}

However, problems will occur if the amount to add $G(e, f)$ is inappropriate. For example if $G(e, f)$ are as high as millions while the counts in the training data is around thousands, the prior counts will dominate the counts in data, that makes the final estimation too flat. Therefore, we need a mechanism to manipulate these quantities rather than merely trust them. We care about two aspects of a function $G$ which affects the final estimation. They are the ratios among the values of the function $G$ and their overall altitude. Scaling all the $G(e, f)$ values by a same number of times is our choice because it can manipulate the overall altitude of the function $G$ while keeping the ratios among values of words which we consider as the most interesting information of $G$. The estimation after that becomes:
\begin{equation}
  t(f \mid e) = \frac{count(f, e) + \lambda G(e, f)}{count(e) + \sum_f \lambda G(e, f)}
\end{equation}

The number of times to scale $\lambda$ is used to adjust the effect of $G(e, f)$ to the estimation. In other words, $\lambda$ is a way to tell how we trust the supplied function. If the $G(e, f)$ is reasonable, $1$ should be assigned $\lambda$ to keep the amount to add unchanged. If the $G(e, f)$ is generally too high, $\lambda$ will take a relative small value as expected. However, if not only $G(e, f)$ are generally high but also the ratios among values of $G(e, f)$ are very unreasonable, the $\lambda$ will be very near $0$.

Any supplied function $G$ can be used in this framework. In the case of Moore's method, we have an interpretation that $G$ is a constant function returning $1$ for every input while $\lambda$ will play the same role as $n$. For other cases, no matter whether the function $G$ is appropriate or not, $\lambda$ will be chosen to appropriately adjust the effect to the final estimation.

The parameter $\lambda$ in our framework is learnt from data to best satisfy some requirements. We will study many sorts of learning from learning on unannotated data to annotated data, from the discrete objective functions to their adaptation to continuous functions. These methods are presented in the next section.

\section{Learning the scaling factors}
\subsection{Approaches to learn the scaling factor}
In this section, several approaches to learn the scaling factor will be presented. We assume the existence of one additional corpus of bilingual pairs $(\textbf{f}, \textbf{e})$ and the word alignments $\bar{\textbf{a}}$ between these pairs which are annotated manually. As the traditional Maximum Likelihood principle, the quantity to maximize is the probability of both $\textbf{f}$ and $\textbf{a}$ given $\textbf{e}$ with respect to $\lambda$.
\begin{align}
  \lambda &= \arg\max_{\lambda} \prod_k p(\textbf{f}^{(k)}, \bar{\textbf{a}}^{(k)} \mid \textbf{e}^{(k)}; \textbf{t}) \\
  &= \arg\max_{\lambda} \prod_k \prod_j t(f^{(k)}_j \mid e^{(k)}_{\bar{a}^{(k)}_j})
\end{align}

where $t(f \mid e)$ is the parameter which is estimated as Expectation Maximization algorithm with $\lambda \; G(e, f)$ is added in the maximization step of each iteration.

However, there is another more popular method based on the error count which is the number of deduced alignment links which are different from links in human annotation. The parameter $\lambda$ in this way is learnt to minimize the total number of the error count.

\begin{align}
  \lambda &= \arg\min_{\lambda} \sum_{k} \sum_{j} [\bar{a}^{(k)}_j \neq {\hat{a}}^{(k)}_j]\\
  &= \arg\min_{\lambda} \sum_{k} \sum_{j} [\bar{a}^{(k)}_j \neq \arg\max_i p(a^{(k)}_j = i)]
\end{align}
for $\bar{a_j}$ is an annotated alignment link and $\hat{a}_j$ is the alignment link produced by the model.

We also experience another method which do not require the additional annotated data. We instead utilize unannotated development data. This additional data is obtained by dividing the original corpus into two parts which are known as the training set and the development set. For a considered $\lambda$ value, the word translation probability is estimated from the training set while the $\lambda$ is later evaluated with the development set. The development set has the same scheme with the training set when no alignment between words is labeled. Therefore, we can only apply the Maximum Likelihood principle. The scaling factor $\lambda$ we desire has the value maximizing the likelihood of the development set as explicitly described below.

\begin{align}
  \lambda &= \arg\max_{\lambda} \prod_k p(\textbf{f}^{(k)} \mid \textbf{e}^{(k)})\\
  &= \arg\max_{\lambda} \prod_k \prod_j \sum_i t(f^{(k)}_j \mid e^{(k)}_i)
\end{align}

All above methods are reasonable. The Maximum Likelihood estimation forces the parameter to respect what we observe. For the case of annotated data, both sentences and annotated alignments have the likelihood to maximize. When there is no annotated validation data, we have to split the training data into two parts and consider one of them as the development data. In this case, only the sentences in the development data have the likelihood to maximize. The method of minimizing the error count is respected to the performance of the model when aligning. It pays attention to how many times the estimated model makes wrong alignments and try to reduce this amount as much as possible.

Although each method deal with different issues, we will not discuss in details about this matter. All these issues make the model better in their own perspective. It would be better to judge them on experiences on data. Instead, we want to compare them in term of computational difficulties. Although it is just the problem of optimizing a function of one variable, there are still many issues to discuss when optimizing algorithms are not always perfect and often have poor performances when dealing with particular situations. In the next section, we analyze the the continuity of the functions and its impacts on the optimizing algorithms.

\subsection{Optimizational aspects of the approaches}
For each parameter of smoothing, we will obtain the corresponded parameter of the model. Because in Expectation Maximization algorithm, each iteration involves only fundamental continuous operators like multiplying, dividing, adding. The function of the most likely parameter of the model for a given parameter of smoothing is also continuous. However, the continuity of the model parameter does not always lead to the continuity of the objective function. This indeed depends much on the nature of the objective function.

The method of minimizing the alignment error count and maximizing the likelihood are quite different in the aspect of continuity when apply optimization techniques. The method of minimizing the alignment error count is discrete due to the $argmax$ operator. The likelihood is continuous due to multiplying only continuous quantities. This means that optimizing with respect to the likelihood is usually easier than that for the alignment error count.

Most optimization algorithms prefer continuous functions. With continuous functions, the algorithms always ensure a local peak. However, with discrete functions, the algorithms do not have such a property. When the parameter changes by an insufficient amount, the discrete objective function does not change. Once the difference of the parameter reaches a threshold, the functions will change by a significant value, at least 1 for the case of the error count function. Therefore there is a significant gap inside the domain of the discrete function at which the optimization algorithms may be trapped. This pitfall usually makes the algorithms to claim a peak inside this gap. The continuous functions instead always change very smoothly corresponding to how much the parameter changes. Therefore, there is no gap inside, that avoid such a pitfall.

For discrete functions, well-known methods for continuous functions are no longer directly applicable. We can actually still apply them with pseudo derivatives by calculating the difference when changing the parameter by a little amount. However, as explained above, the algorithms will treat a point in a gap as a peak when they see a 0 derivative at that point. There are algorithms for minimization without derivatives. Brent algorithm (\cite{brent2013algorithms}) is considered to be an appropriate solution for functions of a single variable as in our case. Although the algorithm do not require the functions to be continuous, the performance will not be good in these cases with the same sort of pitfall caused by gaps.

There is another solution for this problem that we can smooth the objective function. The work \cite{och2003minimum} gives a very useful study case when learning parameters for the phrase-based translation models. Adapting the technique used in that paper, we have our adaptation for the alignment error count, which is an approximation of the original objective function but having the continuous property. The approximation is described as below

\begin{align}
&\sum_{j} [\bar{a}_j \neq {\hat{a}}_j] \\
=& \sum_{j} [\bar{a}_j \neq \arg\max_i p(a_j = i)]\\
=& \sum_{j} (1 - [\bar{a}_j = \arg\max_i p(a_j = i)])\\
\approx & \sum_{j} (1 - \frac{p(a_j = \bar{a}_j)^\alpha}{\sum_i p(a_j = i)^\alpha}) \label{eq:approx-transition}
\end{align}
for a sufficiently high $\alpha$.

When $\alpha$ is getting larger, the quantity $p_i = p(a_j = i)^\alpha$ will be amplified and the differences among $p_i$ will be clearer. When $\alpha$ goes to infinite, the differences will be clearest, that makes the $p_{{\hat{a}}_j}$ to dominate other $p_i$ because $p_{{\hat{a}}_j}$ is the amplification of $p(a_j = {\hat{a}}_j)$, which is the largest of all $p(a_j = i)$. This dominating quantity $p_{{\hat{a}}_j}$ takes almost all of proportion of the sum of all $p_i$. It means that for $i = {\hat{a}}_j$, the quantity $\frac{p(a_j = i)^\alpha}{\sum_i p(a_j = i)^\alpha}$ will nearly equal to 1 while this quantity for other $i$ is 0. Therefore, the quantity $p_i$ is used as an approximation of $[i = \arg\max_i p(a_j = i)]$ as shown in equation \ref{eq:approx-transition} above. That is the main point of the approximation.

This continuous approximation is very close to the original error count. The larger $\alpha$ is, the closer they are. We can arbitrarily set $\alpha$ as much as the power of computation allows. By having this smooth version of the error count, we can prevent the pitfall due to the discrete function but still retain a high correlation with the evaluation of the model performance. 



\section{Experiments}
\subsection{The adding functions to experience}
Most additive strategies are usually based on heuristic methods. In this paper, we will empirically investigate them.

The first additional method we experience is adding a scaled amount of the number of times the word $e$ appears to the count of each pair $(e, f)$. It has the motivation that the same adding amount for every pair may be suitable for only a small set of pairs. When estimating the word translation probability for a very rare word, this adding amount may be too high while for very popular words, this is rather too low. Therefore, we hope that apply a scaled amount of the count of the source word would increase the result. This modifies the estimation as:
\begin{equation}
  t(f \mid e) = \frac{count(f, e) + \lambda \; n_e}{count(e) + \lambda \; n_e |F|}
\end{equation}
where $n_e$ is the number of times word $e$ appear in the corpus.

The other adding method we want to experience is adding a scale of the dice coefficient. This coefficient is a well-known heuristic to estimate the relation between two words $(e, f)$ as:
\begin{equation}
  dice(e, f) = \frac{2 \; count(f, e)}{count(f) + count(e)}
\end{equation}

A scaled amount of the dice coefficient, $\lambda \; dice(e, f)$ is added to the count of pair $(e, f)$ as:
\begin{equation}
  t(f \mid e) = \frac{count(f, e) + \alpha \; dice(e, f)}{count(e) + \alpha \; \sum_{f'} dice(e, f)}
\end{equation}

This function makes the smoothing a bit different from usual smoothing techniques that it does not add any amount to the counts of word pairs that do not co-occur in any sentence pair. Instead, it encourages the parameter of the model to get closer to the dice coefficient. Although word alignment models merely based on dice coefficient is inefficient (\cite{och2003systematic}), adjusting the baseline parameter estimation by an appropriate time of this amount could be an improvement.

\subsection{Performance of the adding functions and the objective functions}
We have experiments on the Europarl corpora (\cite{koehn2005europarl}) of German-English and English-Spanish. We extract 100,000 bilingual sentence pairs from each corpus. For the German-English corpus, we have 150 annotated sentence pairs. For the English-Spanish corpus, the work in \cite{lambert2005guidelines} gives us 500 annotated sentence pairs.

The annotated alignments are very general. There is no requirement that each word in the target sentence is aligned to no more than one word in the source sentence. They are indeed symmetric alignments that one word in the target sentence can be aligned to more than one word in the source sentence and vice versa. Not only so, there are 2 types of annotated alignment links: possible links and sure links. This is due to the fact that word alignments are actually very obfuscated in term of human evaluation. An alignment chosen by one person does not mean that other people will choose it as well. Therefore, there will be two sets of alignment links to be taken into consideration. The first set contains all the possible links chosen by at least one person. The other set instead consists of only sure alignment links which everyone agree on. The set of sure links is of course a subset of the set of possible links.

This sort of annotated data is for Alignment Error Rate (AER) evaluation as in \cite{och2003systematic}. Denote the set of sure links $S$, the set of possible links $P$, and the set of links are decided by the word alignment model $A$, we have an adaptation for the common metric: precision and recall.

\begin{equation}
  Precision = \frac{\mid P \cap A \mid}{\mid A \mid}
\end{equation}

\begin{equation}
  Recall = \frac{\mid S \cap A \mid}{\mid S \mid}
\end{equation}

Instead of F-measure or some other popular metrics derived from precision and recall, AER, a more suitable metric, is commonly used for word alignment problem:

\begin{equation}
  AER = 1 - \frac{\mid P \cap A \mid + \mid S \cap A \mid}{\mid A \mid + \mid S \mid}
\end{equation}

With this metric, a lower score we get, a better word alignment model is.

Because of the speciality of the word alignment problem, the sentence pairs in the testing set are also included in the training set as well. In the usual work-flow of a machine translation system, the IBM models are trained on a corpus, and later align words in the same corpus. There is no need to align words in other corpora. If there is such a task, it could be much better to retrain the models on the new corpora. Therefore, the annotated sentence pairs but not their annotation also appear in the training set. In other words, the training phase is allowed to observe only the testing sentences, and only in the testing phase, the annotation of the test sentences could be seen.

However, for the purpose of the development phase, a small set is randomly extracted from the annotated set for the development set. The number of pairs to cut is 50 out of 150 for the German-English corpus and 100 out of 500 for the English-Spanish corpus. These annotations could be utilized in the training phase, but no longer be used for evaluating in the testing set.

For the development phase, the method utilizing annotated word alignment requires the restricted version of word alignment, not the general, symmetric, sure-possible-mixing ones as in the annotated data. Therefore, we have to develop an adaptation for this sort of annotated data. For simplicity and reliability, we consider sure links only. For target words which do not appear in any link, they are treated to be aligned to ``NULL''. In case a target word appears in more than one links, one arbitrary link of them will be chosen.

Our methods do not force the models to be fit a specific evaluation like AER. Instead of learning the parameter directly respected to AER, we use closer evaluations to the IBM models on the restricted alignment, which appears more natural and easier to manipulate with a clearer theoretical explanation. We also see a high correlation between these evaluations and AER in the experiments. 


IBM Models are trained in the direction from German as source to English as target with the German-English corpus. The direction for the English-Spanish corpus is from English as source to Spanish as target. We apply 10 iterations of IBM model 1 in every experiment. With this baseline method, we obtain the AER score of the baseline models as shown in Table~\ref{tbl:baseline-aer}.

\begin{table}
  \centering
  \begin{tabular}{|c|c|c|}
    \hline
    Corpus & German-English & English-Spanish \\
    \hline
    AER score & 0.431808 & 0.55273 \\
    \hline
  \end{tabular}
  \caption{AER scores of the baseline models on corpora}
  \label{tbl:baseline-aer}
\end{table}

Experiment results of our proposed methods are presented in Table~\ref{tbl:de-en-aer} for the German-English corpus and Table~\ref{tbl:en-es-aer} for the English-Spanish corpus. We experience on scaled amounts of three adding strategies: Moore's method (add one), adding the number of occurrences of the source word (add $n_e$), adding the dice coefficient (add dice) and four objective functions: the likelihood of unannotated data (ML unannot.), the likelihood of annotated data (ML annot.), the error count (err. count) and the smoothed error count (smoothed err. count). We apply every combination of adding strategies and objective functions. There are twelve experiments in total. In each experiment, we record the AER score obtained by running the model on the testing data. 

\begin{table}
  \centering
  \begin{tabular}{|c|c|c|c|c|}
    \hline
    & ML unannot. & ML annot. & err. count & smoothed err. count \\
    \hline
    add one & 0.418029 & 0.405498 & 0.42016 & 0.413596 \\
    \hline
    add $n_e$ & 0.431159 & 0.431808 & 0.437882 & 0.430964 \\
    \hline
    add dice & 0.426721 & 0.427469 & 0.42031 & 0.419854 \\
    \hline
  \end{tabular}
  \caption{AER scores of different methods on the German-English corpus}
  \label{tbl:de-en-aer}
\end{table}

\begin{table}
  \centering
  \begin{tabular}{|c|c|c|c|c|}
    \hline
    & ML unannot. & ML annot. & err. count & smoothed err. count \\
    \hline
    add one & 0.533452 & 0.527508 & 0.50605 & 0.505827 \\
    \hline
    add $n_e$ & 0.556888 & 0.553469 & 0.557111 & 0.553304 \\
    \hline
    add dice & 0.543835 & 0.543395 & 0.535291 & 0.535156 \\
    \hline
  \end{tabular}
  \caption{AER scores of different methods on the English-Spanish corpus}
  \label{tbl:en-es-aer}
\end{table}

The result shows that most of the methods decrease the AER score while only the experiments of adding the number of occurrences of source words increases the AER score relatively lightly. We calculate the difference between AER scores of the new methods and those of the baseline by subtracting the old score by the new scores as in Table~\ref{tbl:de-en-aer-diff} and Table~\ref{tbl:en-es-aer-diff} for respectively the German-English and English-Spanish corpora.

\begin{table}
  \centering
  \begin{tabular}{|c|c|c|c|c|}
    \hline
    & ML unannot. & ML annot. & err. count & smoothed err. count \\
    \hline
    add one & 0.013779 & 0.02631 & 0.011648 & 0.018212 \\
    \hline
    add $n_e$ & 0.000649 & 0 & -0.006074 & 0.000844 \\
    \hline
    add dice & 0.005087 & 0.004339 & 0.011498 & 0.011954 \\
    \hline
  \end{tabular}
  \caption{Decreasement in AER scores of different methods on the German-English corpus}
  \label{tbl:de-en-aer-diff}
\end{table}

\begin{table}
  \centering
  \begin{tabular}{|c|c|c|c|c|}
    \hline
    & ML unannot. & ML annot. & err. count & smoothed err. count \\
    \hline
    add one & 0.019278 & 0.025222 & 0.04668 & 0.046903 \\
    \hline
    add $n_e$ & -0.004158 & -0.000739 & -0.004381 & -0.000574 \\
    \hline
    add dice & 0.008895 & 0.009335 & 0.017439 & 0.017574 \\
    \hline
  \end{tabular}
  \caption{Decreasement in AER scores of different methods on the English-Spanish corpus}
  \label{tbl:en-es-aer-diff}
\end{table}

It is obvious that the Moore's add one method is generally the best, adding dice coefficient is reasonably good, while adding the number of occurrences of source words has a poor performance. This reflects how appropriate the adding methods are.

Most of the performances obtained by the method of adding the number of occurrences of source words are poor. However, due to the mechanism of adjusting the effects by the parameter $\lambda$, in the case the AER scores increase, they increase very slightly by unnoticeable amounts. As expectation, the $\lambda$ coefficient should be adjusted to make the new model as good as the baseline model by setting $\lambda = 0$. However, in experiments, a positive $\lambda$ which is very close to $0$ behaves better in the development set but worse in the testing set. This is due to the fact that the development set and the testing set sometimes mismatch at some points. However, this slight mismatch leads to a very small increment in AER, and could be treated as having the same performance as the baseline. Although the method of adding number of occurrences of source words is poor, the method still have a positive result in the corpus of German-English. With the method of maximum likelihood of unannotated data and the method of minimizing the smoothed error count, it decrease the AER scores. However, because this adding method is inappropriate, the amount of AER decreased is once again unnoticeable. We can conclude that this method of adding lightly affects the performance of the model.

The method of Maximum Likelihood on unannotated data has a positive result and can be comparable to that of Maximum Likelihood of annotated data. Although the Maximum Likelihood of unannotated data is worse than other methods utilizing annotated data in more experiments, the differences are not too significant. Therefore, this method of learning is still a reasonable choice in case of lacking annotated data.

As expectation, optimizing algorithms prefer the smoothed version of the error count. In all experiments, this objective function always gives a better result than the original smoothened version of error counts. The differences in some experiments are considerable, notably in the German-English corpus with the method of adding number of occurrences of source words, the smoothed error count gives a positive result while the original one gives a negative result.

In conclusion, the best choice of the objective function is the smoothed error count and the best choice of the strategies to add is the method of adding one, which is already proposed by Moore. In the case of lacking annotated data, the likelihood of the unannotated data is still worth to consider as the objective function. Although Moore's adding method is still the best so far, with our general framework of adding, we can further experience more adding method with an assurance that it will hardly decrease the performance. Perhaps someday with luck or reasoning, another better adding method will be discovered.






\section{Conclusion}
The Maximum Likelihood estimation is the traditional solution for estimating parameters of word alignment models. However, many works have shown its weakness when estimating with sparse data. Smoothing is usually considered to be a good choice for this case. 
This paper has proposed a general framework which allows customizing the additive amount for each case rather than adding a constant amount as Moore's work. 
Inappropriate adding strategies do not harm the model due to a mechanism of adjusting the effects adding amounts to the estimation. We have demonstrated two additional adding strategies. Although the first method which adds a scaled amount of the number of occurrences of source words is not appropriate, the result of alignment is still nearly unchanged because of the scale factor justification. The second strategy which adds a scale amount of the dice coefficient of the word pair is better than the first one because it decreases the error rate of alignment. Although the Moore's adding strategy still gives the best result among the three strategies, with the framework having no limit in strategy, another better method may be found in the future.

We have analyzed different learning approaches using both unannotated and annotated data. 
The method using unannotated data gives a reasonably positive results in our experiments that means that it could be applied in case of lack of annotated data. 

On the other hand, we also analyzed the affect of the continuity of objective functions. We have shown that the discrete error count makes the optimizing algorithm finding more difficulties than the smoothed error count does. Therefore, we proposed a smooth version of the error count which approximates the original discrete function. The experiments show that this smooth version gives the best result amongst all the methods.
\\
\\
\textbf{Acknowledgements}
\\
This work is supported by the Nafosted project 102.012014.22.

\bibliographystyle{apalike}
\bibliography{smooth-mt-params}
\end{document}